\documentclass[letterpaper, 10pt, conference, compsocconf]{IEEEtran}

\def\IEEEbibitemsep{0pt plus .5pt}

\usepackage[T1]{fontenc}

\usepackage[noadjust]{cite}

\usepackage{graphicx}
\graphicspath{{./figures/}}
\DeclareGraphicsExtensions{.pdf,.jpeg,.png}

\usepackage[cmex10]{amsmath} 
\usepackage{amssymb}  
\usepackage{subdepth}  
\usepackage{mathtools} 

\usepackage{url}
\usepackage{bm}  

\usepackage[protrusion=false]{microtype}  

\usepackage{listings}  
\usepackage[straightquotes,zerostyle=d, scaled=0.98]{newtxtt} 

\usepackage[caption=false,font=footnotesize]{subfig}

\usepackage{physics}  
\usepackage{siunitx}
\usepackage{amsfonts}
\usepackage{cancel}  

\usepackage{tabu, booktabs}  
\usepackage{collcell}  
\usepackage{multirow}  
\usepackage{diagbox}   
\usepackage{makecell}  
\usepackage{threeparttablex}  
\usepackage[table, svgnames]{xcolor}   
\usepackage{multicol}  

\makeatletter
\AtBeginEnvironment{lstlisting}{%
\long\def\@makecaption#1#2{%
\ifx\@captype\@IEEEtablestring%
\footnotesize\bgroup\par\centering\@IEEEtabletopskipstrut{\normalfont\footnotesize #1}\\{\normalfont\footnotesize\scshape #2}\par\addvspace{0.5\baselineskip}\egroup%
\@IEEEtablecaptionsepspace
\else
\@IEEEfigurecaptionsepspace
\setbox\@tempboxa\hbox{\normalfont\footnotesize {#1.}\nobreakspace\nobreakspace #2}%
\ifdim \wd\@tempboxa >\hsize%
\setbox\@tempboxa\hbox{\normalfont\footnotesize {#1.}\nobreakspace\nobreakspace}%
\parbox[b]{\hsize}{\normalfont\footnotesize\noindent\unhbox\@tempboxa#2}\vskip\belowcaptionskip\relax%
\else%
\ifCLASSOPTIONconference \hbox to\hsize{\normalfont\footnotesize\hfil\box\@tempboxa\hfil}\vskip\belowcaptionskip\relax
\else \hbox to\hsize{\normalfont\footnotesize\box\@tempboxa\hfil}\medskip%
\fi
\fi
\fi}}
\makeatother

\usepackage[pagebackref=false,breaklinks=true,bookmarks=false,
colorlinks,
citecolor=Green,
linkcolor=red,
urlcolor=NavyBlue,
]{hyperref}

\usepackage[capitalise]{cleveref}
\crefname{equation}{}{}   
\crefname{figure}{Fig.}{Figs.}

\newenvironment{nospaceflalign}
 {\setlength{\abovedisplayskip}{0pt}\setlength{\belowdisplayskip}{0pt}%
 \vspace{-\baselineskip}\setlength{\jot}{0pt}%
  \csname flalign\endcsname}
 {\csname endflalign\endcsname\ignorespacesafterend}

\newcommand{\ourlib}{{wave\_geometry}}

\newcolumntype{Y}{>{\centering\arraybackslash}X}  
\newcolumntype{H}{>{\collectcell\lstinline}l<{\endcollectcell}}  
\newcolumntype{N}{>{$}c<{$}}

\newcommand{\liegrp}[2]{\mathrm{#1}(#2)}
\newcommand{\liealg}[2]{\mathfrak{#1}(#2)}
\newcommand{\SO}[1]{\liegrp{SO}{#1}}
\newcommand{\SE}[1]{\liegrp{SE}{#1}}
\newcommand{\so}[1]{\liealg{so}{#1}}
\newcommand{\se}[1]{\liealg{se}{#1}}

\newcommand{\R}[1]{\mathbb{R}^{#1}}

\newcommand{\Frame}[1]{\mathcal F_{#1}}

\renewcommand{\v}[1]{\mathbf{#1}}
\newcommand {\f}[3]{{}_{#2} \mathbf{#1}_{#3}}
\newcommand {\F}[2]{\mathbf{#1}_{#2}}

\newcommand{\elvec}{p}
\newcommand {\fvec}[2]{\f{\elvec}{#1}{#2}}
\newcommand {\vvec}{\v{\elvec}}
\newcommand{\elvecg}{v}
\newcommand {\fvecg}[2]{\f{\elvecg}{#1}{#2}}
\newcommand {\vvecg}{\v{\elvecg}}
\newcommand{\elrot}{C}
\newcommand {\frot}[1]{\F{\elrot}{#1}}
\newcommand {\vrot}{\v{\elrot}}
\newcommand{\eltf}{T}
\newcommand {\ftf}[1]{\F{\eltf}{#1}}
\newcommand {\vtf}{\v{\eltf}}
\newcommand{\elgrp}{\Phi}
\newcommand {\fgrp}[1]{\F{\elgrp}{#1}}
\newcommand {\vgrp}{\v{\elgrp}}
\newcommand{\elalg}{\varphi}
\newcommand {\falg}[2]{\f{\elalg}{#1}{#2}}
\newcommand {\valg}{\v{\elalg}}

\newcommand{\hatop}{\times}
\newcommand{\hatopname}{{cross}}

\newcommand{\filledcell}{\cellcolor{green!10}}
\newcommand{\checkedcell}{\cellcolor{green!25}\checkmark}
\newcommand{\emptycell}{\cellcolor{black!5}}

\newcommand{\FA}[0]{A}
\newcommand{\FB}[0]{B}
\newcommand{\FC}[0]{C}
\newcommand{\FD}[0]{D}

\begin{document}
%
\title{Manifold Geometry with Fast Automatic Derivatives\\ and Coordinate Frame Semantics Checking in C++}

\author{\IEEEauthorblockN{Leonid Koppel, Steven L. Waslander}
\IEEEauthorblockA{Department of Mechanical and Mechatronics Engineering\\
University of Waterloo\\
Waterloo, Canada\\
\{lkoppel, stevenw\}@uwaterloo.ca}
}



\makeatletter
\def\ps@IEEEtitlepagestyle{
  \def\@oddfoot{\mycopyrightnotice}
  \def\@evenfoot{}
}
\def\mycopyrightnotice{
  {\footnotesize
  \begin{minipage}{\textwidth}
  \copyright~2018 IEEE. Personal use of this material is permitted. Permission from IEEE must be obtained for all other uses, in any current or future media, including reprinting/republishing this material for advertising or promotional purposes, creating new collective works, for resale or redistribution to servers or lists, or reuse of any copyrighted component of this work in other works.
  \end{minipage}
  }
}

\maketitle

\begin{abstract}
Computer vision and robotics problems often require representation and estimation of poses on the SE(3) manifold.
Developers of algorithms that must run in real time face several time-consuming programming tasks, including deriving and computing analytic derivatives and avoiding mathematical errors when handling poses in multiple coordinate frames.
To support rapid and error-free development, we present wave\_geometry, a C++ manifold geometry library with two key contributions: expression template-based automatic differentiation and compile-time enforcement of coordinate frame semantics. We contrast the library with existing open source packages and show that it can evaluate Jacobians in forward and reverse mode with little to no runtime overhead compared to hand-coded derivatives. The library is available at \url{https://github.com/wavelab/wave_geometry}.
\end{abstract}


%
\IEEEpeerreviewmaketitle

\section{Introduction} \label{intro}

Representation and estimation of poses is central to a broad range of computer vision and robotics problems. Most approaches rely on some form of optimization over possible poses, which requires operations over the differentiable manifold of SE(3)~\cite{barfoot2017state}, and frequently leads to long chains of transformations between coordinates frames. This results in a persistent need for the computation of Jacobians for complex expressions of manifold elements, a process that is both time-consuming and error-prone.

Contemporary real-time implementations typically use hand-coded, analytically derived Jacobians and rely on external C++ libraries, examined in \cref{sec:related}, for numerical and optimization routines. For example, Eigen~\cite{eigenweb} may be used for matrix operations, sometimes with a specialized library for manifold geometry. These libraries are typically used with a separate nonlinear least squares solver, such as Ceres~\cite{ceres-solver}. Other optimization frameworks, such as GTSAM~\cite{Dellaert2012}, provide their own manifold representations, as well as handwritten code for the derivatives of common cost functions.

Automatic differentiation (AD) presents the possibility of calculating accurate derivatives of arbitrary functions without an analytical expression~\cite{hoffmann2016hitchhiker}.
While general-purpose AD tools are widely available~\cite{baydin2015automatic}, conventional tools do not readily support differential calculus on manifolds~\cite{Sommer2013}. Recent works extend AD to differentiable manifolds~\cite{hoffmann2016hitchhiker, Sommer2013}, and it is available in GTSAM.
Still, these tools impose considerable runtime overhead compared to handwritten code.

A separate challenge is posed by calculations involving multiple coordinate frames. Ambiguities often lead to logic errors and lost development time~\cite{furgale2014pose, DeLaet2013a}. Libraries which check the semantics of rigid body calculations exist~\cite{DeLaet2013b}, but they perform checks at runtime, also imposing overhead.

This work presents two contributions. First, we use expression templates to implement forward- and reverse-mode automatic differentiation of geometric expressions with significantly less runtime overhead than existing works. Second, we introduce a method for checking the semantic correctness of geometric calculations at compile time, using a system of rules for coordinate frame semantics.
 
Our C++11 implementation, \ourlib{}, is publicly available as an open-source library.\footnote{\url{https://github.com/wavelab/wave_geometry}}

\section{Notation and Preliminaries} \label{notation}
In this document, $\Frame A$ represents a coordinate frame whose origin is the point $A$. $\fvecg{A}{BC}$ denotes a vector quantity of $\Frame C$ relative to $\Frame B$, expressed in $\Frame A$: For example, $\f p{A}{AB}$ is the displacement from $A$ to $B$, expressed in $\Frame A$, $\f \omega{A}{BC}$ is the angular velocity of $\Frame C$ relative to $\Frame B$, measured in $\Frame A$. A rotation between $\Frame A$ and $\Frame B$ is denoted by $\frot{AB}$, and a rigid transformation between $\Frame A$ and $\Frame B$ by $\ftf{AB}$, such that
\begin{align}
\frot{AB}(\fvec{B}{BC}) = \fvec{A}{BC}, \\
\ftf{AB}(\fvec{B}{BC}) = \fvec{A}{AC}.
\end{align}

Rotations and poses present a computational challenge because they do not form a vector space. Instead, they form a non-Euclidean manifold, which only locally resembles Euclidean space. In this work we focus on the special orthogonal group, $\SO3$, and the special Euclidean group, $\SE3$. The informal summary here is based on~\cite{barfoot2017state, Bloesch2016}.

$\SO3$ is the Lie group comprising all valid rotation matrices. Its tangent space is described by its associated Lie algebra, $\so3$, whose elements are skew-symmetric matrices. Elements of $\R3$ can be bijectively mapped to $\so3$ using the \hatopname{} operator
\begin{align}
\vvec ^{\hatop} = 
\begin{bmatrix}
p_1 \\ p_2 \\ p_3\\
\end{bmatrix} ^{\hatop}
= \begin{bmatrix}
0 & -p_3 & p_2\\
p_3 & 0 & -p_1 \\
-p_2 & p_1 & 0
\end{bmatrix}.
\end{align}

Note that in matrix form, $\so3$ is overparametrized, and library implementations use the compact $\R3$ representation. Rotation matrices on $\SO3$ are also overparametrized, but storing a minimal representation leads to singularities. The solution outlined in~\cite{Hertzberg2013} is to store overparametrized states while updating them via small perturbations, represented minimally in the tangent space.

Elements of $\so3$ are related to $\SO3$ by the exponential map, $\exp : \so3 \rightarrow \SO3$. For small rotations, this map is bijective, and its inverse is the logarithmic map, $\log : \SO3 \rightarrow \so3$.

Corresponding mappings can be defined for $\SE3$, the group of proper rigid transformations, and its Lie algebra $\se3$. These mappings are used to define nonlinear addition and subtraction operators, called \emph{boxplus} and \emph{boxminus}. One definition is~\cite{Bloesch2016}
\begin{align} 
\vgrp \boxplus \valg &= \exp(\valg)\circ\vgrp   \label{eq:boxplus},  \\
\vgrp_1 \boxminus \vgrp_2 &= \log(\vgrp_1 \circ \vgrp_2^{-1}) \label{eq:boxminus}
\end{align}
where $\vgrp \in \SO3,\; \valg \in \so3$ or $\vgrp \in \SE3,\; \valg \in \se3$.
This shorthand notation is used throughout this document, while $\vvecg$ represents any element of $\R3$, $\so3$, or  $\se3$.

The $\boxplus$ and $\boxminus$ operators let us define derivatives of manifold operations with respect to small changes in the tangent space and calculate Jacobians with the dimensions of the minimal representation. These \textit{local Jacobians} are more useful for optimization and more efficient to compute than the \textit{global Jacobians} produced by conventional AD, which are tied to particular overparametrized representations~\cite{Sommer2013}.

\Cref{tab:ops} summarizes these operations, which are supported by \ourlib{}.

\begin{table}[tb] 
\caption{Summary of Basic Manifold Operations}
\label{tab:ops}
\setlength{\tabcolsep}{0pt} 
\begin{tabu} to \columnwidth { l *{3}{X[c$] } } 
\toprule 
 & \SO3,\, \SE3 & \so3,\, \se3 & \R3 \\
  & \vgrp & \valg & \vvec
\\
\midrule
  unary &   \vgrp^{-1},\; \log(\vgrp)  &  -\varphi,\; \exp(\varphi) & -\vvec \\
  $\SO3,\, \SE3$\hspace{1em} &  \vgrp\circ\vgrp,\; \vgrp\boxminus\vgrp &  \vgrp \boxplus \varphi &   \vgrp \cdot \vvec \\
  $\so3,\, \se3$ &  & \varphi \pm \varphi &  \varphi \cdot \vvec \\
  $\R3$ &  &  &  \vvec \pm \vvec \\
\bottomrule
\end{tabu}
\end{table}

\section{Related Work} \label{sec:related}

\begin{table}[tb]
\begin{ThreePartTable}
\caption{Comparison of C++ Geometry Libraries}
\label{tab:comparison}
\setlength{\tabcolsep}{1pt} 

\begin{tabu} to \columnwidth{  X[1, l]  *{6}{X[c]} }
\toprule
 Library & Manifold ops. & Manifold Jacobians & Any scalar & Maps & Expr. Jacobians & Frame checking \\
\midrule
Eigen & \emptycell & \emptycell & \checkedcell & \checkedcell & \emptycell & \emptycell  \\
KDL & \filledcell * & \filledcell * & \emptycell & \emptycell & \emptycell & \emptycell \\
MTK & \checkedcell & \emptycell  & \checkedcell & \emptycell & \emptycell & \emptycell  \\
Sophus & \checkedcell & \emptycell & \checkedcell & \checkedcell & \emptycell & \emptycell  \\
Kindr & \checkedcell &  \filledcell *  & \checkedcell & \emptycell & \emptycell & \emptycell  \\
MRPT & \checkedcell & \checkedcell & \emptycell & \emptycell & \emptycell  & \emptycell  \\
g2o & \checkedcell & \checkedcell & \emptycell & \emptycell & \emptycell & \emptycell  \\
GTSAM & \checkedcell & \checkedcell & \emptycell & \emptycell & \checkedcell & \emptycell  \\
This~work & \checkedcell & \checkedcell & \checkedcell & \checkedcell & \checkedcell & \checkedcell  \\
\bottomrule
\end{tabu}
\begin{tablenotes}
\item *Partial
\end{tablenotes}
\end{ThreePartTable}
\end{table}

Automatic differentiation encompasses a wide body of work in the computer science and machine learning fields~\cite{Carpenter2015, baydin2015automatic}, and existing tools such as ADOL-C~\cite{walther2009getting}, Adept~\cite{Hogan2014}, and Stan Math~\cite{Carpenter2015} provide automatic differentiation of arbitrary functions. While we do not intend to compete with these advanced, general-purpose libraries, we apply some of their techniques, such as expression templates~\cite{Aubert2001, Phipps2012} to the specific domain of pose estimation on manifolds.

In the fields of computer vision and robotics, Eigen~\cite{eigenweb} is widely used for storage and manipulation of states. Its Geometry module provides transformations and matrix and quaternion parametrizations of rotations. Though it lacks manifold operations, Eigen is ubiquitous, used internally in some form by every other library listed here.

Ceres Solver~\cite{ceres-solver} is a nonlinear least squares solver which optionally performs AD. Though it does not provide geometric types, it supports on-manifold optimization through its \lstinline{LocalParameterization}. Ceres' interface uses raw arrays, which can be interpreted as matrices using Eigen's \lstinline{Map} class. For example, the OKVIS visual-inertial odometry package~\cite{leutenegger2015keyframe} uses Ceres, Eigen, and hand-coded analytic Jacobians.

The Manifold Toolkit (MTK)\footnote{\url{http://openslam.org/MTK.html}}~\cite{Hertzberg2013}, Kindr\footnote{\url{https://github.com/ethz-asl/kindr}}, and Sophus\footnote{\url{https://github.com/strasdat/Sophus}} implement manifold operations, though they do not provide Jacobians for all operations in \cref{tab:ops}.

The Kinematics and Dynamics Library (KDL)\footnote{\url{http://www.orocos.org/kdl}} provides geometric classes, focusing on kinematic chains. While it does not explicitly provide manifold operations, it supports pose interpolation and Jacobians of kinematic chains.

The Mobile Robot Programming Toolkit (MRPT)\footnote{\url{https://www.mrpt.org}} is a collection of libraries for robotics applications including SLAM, computer vision, and motion planning. It provides a 3D geometry library, including Jacobians for operations on~$\SE3$.

GTSAM~\cite{Dellaert2012} and g2o~\cite{kummerle2011g} are frameworks for nonlinear optimization based on factor graphs. They include their own implementations of manifold geometry, including Jacobians. Notably, GTSAM 4.0\footnote{\url{https://bitbucket.org/gtborg/gtsam}} includes automatic differentiation of arbitrary expressions. GTSAM's implementation differs from ours in that it uses runtime polymorphism, not expression templates, to build expression trees. Differences between the two approaches are discussed in \cref{results}.

Many of the ideas in this work are described in~\cite{Sommer2013}, which defines \textit{block automatic differentiation} on differentiable manifolds and its application to robotics problems. We discuss this approach in \cref{ad}. We evaluate our implementation on a similar example to~\cite{Sommer2013} and improve on their results in terms of computational efficiency.

\Cref{tab:comparison} presents a comparison of commonly used open-source C++ libraries. This comparison is narrowly focused on the features we are targeting---of course, these mature libraries have a wide range of other features and use cases. Here, \textit{any scalar} refers to support for numeric types other than \texttt{float} and \texttt{double}, which makes functions compatible with many general-purpose AD libraries. Among the libraries listed, \textit{any scalar} support also indicates a \textit{header-only} library. While header-only libraries often increase compile time compared to precompiled libraries, they are more flexible and can produce more highly optimized code.

\textit{Maps}, sometimes called \textit{views}, allow zero-overhead reuse of raw memory buffers, as implemented in Eigen's \lstinline{Map} class, and allows efficient interfacing with third-party libraries~\cite{eigenweb}.

\textit{Frame checking} refers to the use of software to systematically check geometric relations. This concept is discussed in~\cite{DeLaet2013a, DeLaet2013b}, which lists common errors, introduces \textit{semantics checking}, and implements it for KDL. Our method, presented in \cref{frames}, is inspired by long-existing methods for dimensional analysis in C++~\cite{barton1994scientific}.

\section{Expression Templates} \label{expr-templates}

\begin{figure}[tb]
\centering
\includegraphics{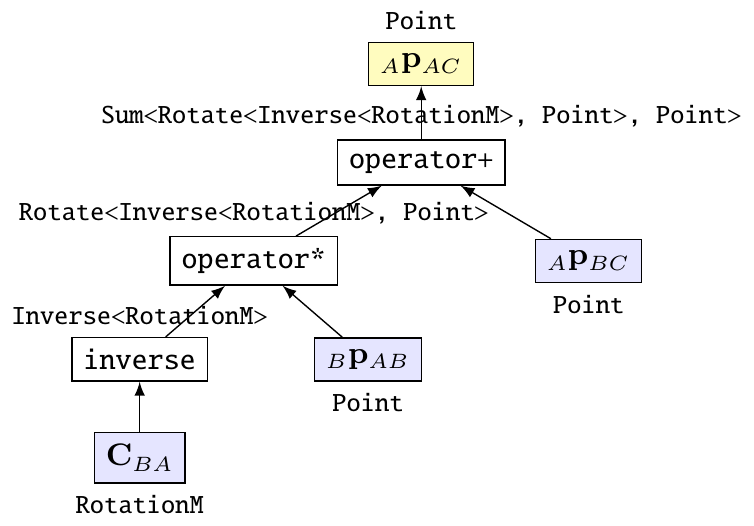}
\caption{Expression tree showing propagation of types at compile time for~\cref{eq:expr1}.
Each return type encodes the structure of the subtree below.}
\label{fig:expr1}
\end{figure}

Expression templates (ET) are a C++ programming technique originally invented to optimize numeric array operations~\cite{Vandevoorde2002templates}. Their defining feature is encoding mathematical expressions as template arguments. Using ET, the C++ expression \verb|C_BA.inverse() * B_p_AB + A_p_B|, representing
\begin{equation} \label{eq:expr1}
{\frot{BA}}^{-1} \fvec{B}{AB} + \fvec{A}{BC}
\end{equation}
does not return a \texttt{Point} object. Instead, each function returns an expression object, producing the tree shown in~\cref{fig:expr1}. The three objects in~\cref{eq:expr1} become leaf nodes, connected by unary or binary operations hold references to one or two children. The the last operation is the root node. In this case, the return type of \verb|operator+| encodes the entire structure of~\cref{eq:expr1}. When it is assigned to a \texttt{Point} object, the expression tree is evaluated.

While early ETs served to eliminate unnecessary temporaries and loops, they relied on the compiler's decision-making to produce efficient low-level code. Modern ``smart'' ET implementations use ETs primarily as a parsing mechanism and apply their own optimizations, such as choosing the order of evaluation of subexpressions~\cite{iglberger2012expression}. 

Interestingly, while Eigen makes extensive use of ET and most of the geometry libraries in \cref{tab:comparison} use Eigen for internal representation, none of them---including Eigen's own geometry classes---use ET in their own interfaces.

As motivating examples, consider the following pair of equations and their code representations:
  \begin{alignat}{2}
    & \fvec{}{2} = \frot{1}  \frot{2} \fvec{}{1} \qquad&& \texttt{p2 = C1 * C2 * p1;}
    \label{eq:ex1} \\
    & \fvec{}{2} = \vtf^{-1} \fvec{}{1}  && \texttt{p2 = T.inverse() * p1;}
    \label{eq:ex2}
  \end{alignat}

Without expression templates, \cref{eq:ex1} is evaluated left-to-right, performing a matrix-matrix product. It is substantially cheaper to move right-to-left and perform only matrix-vector products~\cite{iglberger2012expression}.  In \cref{eq:ex2}, the actual inverse is not needed; it is cheaper to evaluate the transformation in one step using modified coefficients of $\vtf$. Without expression templates,  such optimizations require changes to user code. For example, KDL warns users to insert parentheses, \lstinline{R1*(R2*p1)}, for~\cref{eq:ex1}, and to call a separate function, \lstinline{T.Inverse(p1)}, for~\cref{eq:ex2}.

\subsection{Implementation}\label{expr-impl}

We implement ET by applying the curiously recurring template pattern (CRTP), a technique for code reuse and compile-time polymorphism also used by Eigen. CRTP's application to ET is described by~\cite{Hardtlein2010}. Following Eigen,\footnote{Eigen's internal implementation is described at \url{http://eigen.tuxfamily.org/index.php?title=Working_notes_-_Expression_evaluator}} we traverse expression trees using a recursive \emph{evaluator}, which caches intermediate results for use by subsequent operations.

Each intermediate value is not necessarily a dense matrix object: it can be an Eigen expression, allowing lazy evaluation. We further optimize by extending Eigen with several frequently recurring expressions. For example, our \texttt{Identity} expression represents an $N\times N$ identity matrix which is trivially eliminated from products. Our \texttt{CrossMatrix} expression wraps a vector $\v a \in \R3$, lazily evaluating to $\v a^\hatop \in \R{3\times3}$ but producing efficient cross product code when multiplied.

\section{Automatic Differentiation} \label{ad}

\begin{figure*}
\centering
\subfloat[Forward evaluator. Since two variables have the same type, they must be checked at runtime.
]{
\includegraphics{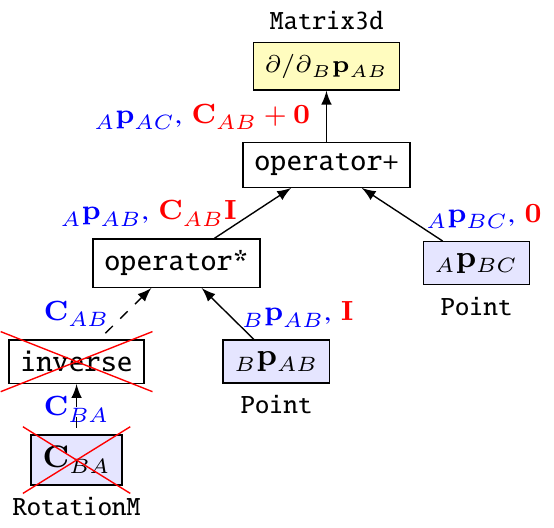}
\label{fig:treea}
} \hspace{1cm}\subfloat[Strongly typed forward evaluator. Since all leaves have different types, the calculation is fully predetermined and no branching occurs.]{
\includegraphics{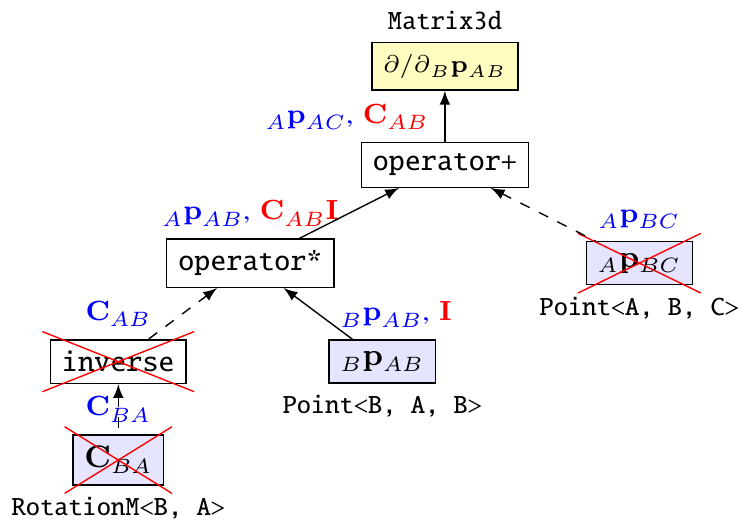}
\label{fig:treeb}
} \hspace{2mm} \subfloat[
Strongly typed reverse evaluator. Jacobians for all leaves are calculated in one sweep.
]{
\includegraphics{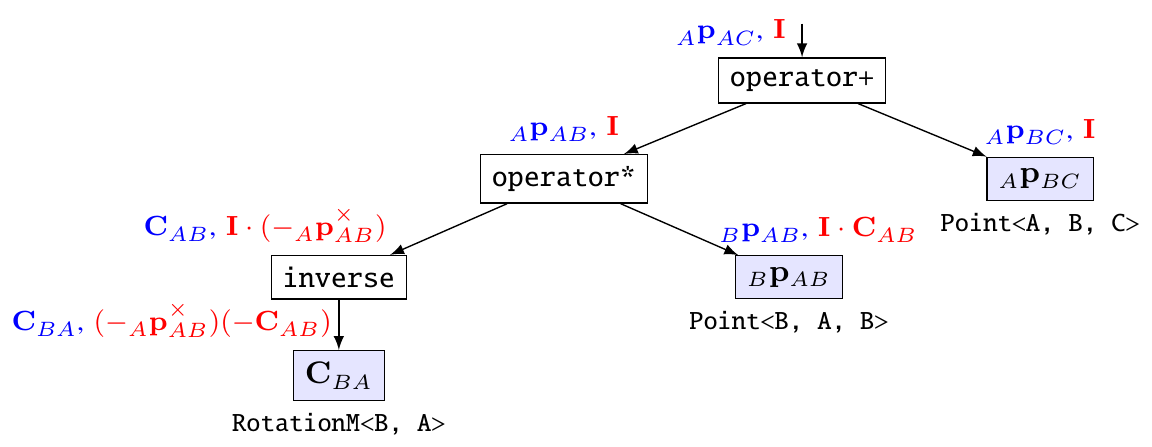}
\label{fig:treec}
\hspace{1.8cm} 
}
\caption{Computational graphs showing differentiation of example \cref{eq:expr1}, comparing Forward (a),  Typed Forward (b), and Reverse (c) implementations.
Cached values from the original function evaluation are shown in blue, and derivatives in red. Crossed-out nodes are not traversed at runtime.}
\label{fig:trees}
\end{figure*}

In this section, we explore three AD implementations built on the expression template scaffolding.
We first clarify our use of the term ``automatic differentiation'': while many works define AD in terms of functions of dual numbers~\cite{hoffmann2016hitchhiker}, it can refer to any technique that applies symbolic derivatives of elementary operations to propagate numerical results during execution~\cite{baydin2015automatic}. This section assumes a basic familiarity with the propagation of gradients and adjoints using the chain rule, as described in~\cite{griewank2008evaluating, baydin2015automatic, hoffmann2016hitchhiker}.

Three possible groupings of AD are forward vs. reverse mode, operator overloading vs. source code transformation, and scalar vs. matrix-valued operations~\cite{Andersson2012}. In this work, we describe implementations of both forward- and reverse-mode algorithms. We use operator overloading, propagating derivative through the expression graph of the original function. Unlike general-purpose AD libraries which treat matrices as multivariate containers of scalar variables~\cite{Carpenter2015}, we work with matrix-valued elementary operations, an approach called block automatic differentiation~\cite{Sommer2013}.

\subsection{Forward-Mode AD}\label{ad-forward}
The forward-mode evaluator, illustrated in \cref{fig:treea}, closely resembles the standard \verb|Evaluator| (see \cref{expr-impl}), with the addition of the chain rule. However, instead of traversing the original expression tree, the Jacobian evaluator traverses the \verb|Evaluator| tree. This design allows reuse of the \verb|Evaluator|'s cached values: for each operand of an elementary operation, the Jacobian computation function is given the values of the operation and of its operands.

We use type information encoded in the expression tree to eliminate unnecessary operations. Evaluating the Jacobian of a leaf node with respect to a target variable requires an identity check. If the two have different types, the Jacobian is known to be zero at compile time, and we avoid traversing that branch of the tree.

If two variables in an expression have the same type, however, it is impossible to predict the evaluation flow (since we cannot tell at comile time whether the two inputs alias a single object~\cite{Hogan2014}), eliminating a host of possible optimizations. Since we cannot predict return types, intermediate values must be dense matrices, not arbitrary expressions. We mitigate this cost somewhat by wrapping intermediate Jacobian matrices in \texttt{boost::optional}, which allows efficient return of ``zero".

\subsection{Strongly Typed Forward-Mode AD}\label{typed-ad}

Still, the implementation can be more efficient if it is guaranteed that all instances of a type refer to the same object. That is the principle behind our second forward AD implementation, illustrated in \cref{fig:treeb}. Normally, the uniqueness guarantee covers only a small subset of expressions, unless variables are ``tagged'' specifically to differentiate their types. However, the coordinate frame semantics system presented in \cref{frames} conveniently has the same effect, extending the guarantee to a large set of physically meaningful expressions. For expressions which do make repeated use of one variable, the typed evaluator can be invoked manually, with the uniqueness assumption checked at runtime in debug builds.

\subsection{Reverse-Mode AD}\label{reverse-ad}
In reverse mode, illustrated in \cref{fig:treec}, the derivatives of one output with respect to all inputs are calculated by backwards application of the chain rule. For geometric expressions with one output, this mode calculates all Jacobians in one sweep. Like the previous evaluator, it relies on the uniqueness guarantee to generate optimized code.

\section{Coordinate Frame Semantics Checking} \label{frames}
Geometric expressions encode \emph{semantics}: information about the meaning of the symbols and their relationships. This section gives an overview of the concept, which is discussed in detail in~\cite{DeLaet2013a, DeLaet2013b}; describes how it can prevent common coding mistakes at compile time; and presents a system of rules for semantics which extends manifold operations.

\subsection{Expressing Semantics Through Notation}

Consider transforming the position of a landmark $L$, from a camera frame, $\Frame C$, to a robot body frame, $\Frame B$. We denote the landmark's position unambiguously as $\fvec{C}{CL}$ (``the vector from $\Frame C$ to $\Frame L$, expressed in  $\Frame C$''), although it is common to omit the suffixes and write $\fvec{C}{}$. One possible expression for the position in the body frame is
\begin{equation} 
\label{eq:framexample}
\fvec{B}{BL} = (\F{C}{CB})^{-1} \fvec{C}{CL} + \fvec{B}{BC}.
\end{equation}
The subscript notation recommended by Furgale~\cite{furgale2014pose} explicitly shows the semantics of each symbol. Crucially, the semantics of the final result are wholly determined by combinations of elementary operations:
\begin{align}
(\F{C}{CB})^{-1} \fvec{C}{CL} + \fvec{B}{BC}
&= \F{C}{BC} \fvec{C}{CL} + \fvec{B}{BC} \\ 
&=\fvec{B}{CL} + \fvec{B}{BC} \\
&= \fvec{B}{BL}.
\end{align}
Mistakes such as an omitted inverse result in invalid operations, which are indicated by mismatching subscripts:
\begin{equation} \label{eq:framemistake}
\fvec{B}{BL} = \F{C}{C\color{red}B} \fvec{\color{red}C}{CL} + \fvec{B}{BC}.
\end{equation}
This analysis is independent of the numeric value and parametrization of the variables. Equation~\cref{eq:framemistake} is semantically incorrect despite being readily computable.

\subsection{Semantics in Code}

To help avoid mistakes such as~\cref{eq:framemistake} in code, Furgale~\cite{furgale2014pose} recommends using prefix notation in variable names, as in \cref{lst:frame_names}.
While this convention helps programmers parse the code and spot mistakes, it does not prevent an invalid expression from compiling. To do so, we encode coordinate frame semantics into the \textit{types} of variables, instead of their names, as in \cref{lst:frames}.

\begin{lstlisting}[caption={Frame semantics expressed in variable names}, label={lst:frame_names}]
Point B_p_B_L = C_C_B.inverse() * C_p_CL + B_p_B_C;
\end{lstlisting}

\begin{lstlisting}[caption={Frame semantics embedded in \ourlib{} types}, label={lst:frames}]
struct Body;       // Represents the robot frame
struct Camera;     // Represents the camera frame
struct Landmark;  // Represents a landmark frame

Point<Body, Body, Landmark> landmarkToBody(
  const Rotation<Body, Camera>& R_cam,
  const Point<Body, Body, Camera>& p_cam,
  const Point<Camera, Camera, Landmark>& p) {
    return R_cam * p + p_cam;
}
\end{lstlisting}

This approach states the meaning of each variable and function at declaration, without cluttering internal code. Invalid operations will cause an error at compile time.

Frame checking is integrated into \ourlib{}'s expression template implementation. Each expression's semantics are encoded in a traits class, and each expression defines rules for its inputs. C++'s \texttt{static\_assert} mechanism is used to trigger a compilation error, printing an explanation if the rules are broken. The coordinate frame template arguments are arbitrary type names, which can be declared as \texttt{struct}s with no definition. This approach has no runtime overhead, in contrast to~\cite{DeLaet2013b}, which stores semantic information inside objects and performs checks at runtime.

\subsection{Rules for Semantics Checking}
\Cref{tab:rules} presents a system of rules used for checking and propagating coordinate frame semantics. To formulate this system, we first define the minimal set of coordinate frame descriptors needed to fully describe each expression. Vector quantities need three descriptors, and coordinate mappings need two~\cite{furgale2014pose}. Note that we use a smaller set of descriptors than~\cite{DeLaet2013a}, because we do not model rigid bodies. We then choose a set of rules which is internally consistent under vector space and manifold axioms~\cite[eq. (11)]{Hertzberg2013}.

\begin{table}[tb]
\begin{ThreePartTable}
\caption{Rules for Semantics of Geometric Operations}
\label{tab:rules}
\begin{subequations}
\begin{nospaceflalign}
\toprule
& \text{Operation} & & \text{Rule}  & \nonumber \\
\midrule
& \text{Sum} & \fvecg{\FD}{\FA\FC}  &= \fvecg{\FD}{\FA\FB} + \fvecg{\FD}{\FB\FC} & \\[1.5pt]
& & &= \fvecg{\FD}{\FB\FC} + \fvecg{\FD}{\FA\FB} & \nonumber \\
\midrule
& \text{Negative} &\fvecg{\FD}{\FB\FA}  &=   - \fvecg{\FD}{\FA\FB} & \\
\midrule
& \text{Difference} & \fvecg{\FD}{\FA\FB}  &=  \fvecg{\FD}{\FA\FC}  - \fvecg{\FD}{\FB\FC}  & \\ 
\midrule
& \text{Scaling} &\fvecg{\FA}{\FB\FC}  &= a(\fvecg{\FA}{\FB\FC}),\quad a \in \R{} & \\
\midrule
& \text{Composition} & \fgrp{\FA\FC}  &=  \fgrp{\FA\FB} \circ \fgrp{\FB\FC}  & \\
\midrule
& \text{Inverse} &  \fgrp{\FB\FA}  &= (\fgrp{\FA\FB})^{-1}  & \\
\midrule
& \text{Rotation} & \fvec{\FD}{\FB\FC}  &=  \frot{\FD\FA} (\fvec{\FA}{\FB\FC})  & \\
\midrule
& \text{Transformation} &  \fvec{\FA}{\FA\FC} &= \ftf{\FA\FB} (\fvec{\FB}{\FB\FC}) & \\[1.5pt]
& & &=  \frot{\FA\FB} (\fvec{\FB}{\FB\FC}) + \fvec{\FA}{\FA\FB} & \nonumber \\
\midrule
& \text{Manifold plus} & \fgrp{\FA\FB} &= \fgrp{\FA\FB} \boxplus \falg{\FA}{\FA\FB}  & \label{eq:rule-boxplus} \\
\midrule
& \text{Manifold minus} &  \falg{\FA}{\FA\FB} &= \fgrp{\FA\FB} \boxminus \fgrp{\FA\FB}  & \label{eq:rule-boxminus}\\
\midrule
& \text{Exp map*} &  \fgrp{\FA\FA} &= \exp(\falg{\FA}{\FA\FB}) & \label{eq:rule-expmap} \\
\midrule
& \text{Log map*} &  \falg{\FA}{\FA B} &= \log_B(\fgrp{\FA\FA}) & \label{eq:rule-logmap} \\
\bottomrule \nonumber
\end{nospaceflalign}
\end{subequations}
\begin{tablenotes} \vspace{-\baselineskip}
\item Valid operands for each operation are shown on the right hand side, and the result is shown on the left. Repeated frame labels must match.
\item *See text for discussion.%
\end{tablenotes}%
\end{ThreePartTable}%
\end{table}

While semantics checking can catch common mistakes, it is important to realize it cannot handle every case. For example, it cannot verify that the numeric value of $\fvec{A}{BB}$ is zero. Indeed, $\fvec{A}{BB}$ could reasonably represent a residual obtained by subtracting a measured and estimated vector, or a perturbation to be added to $\fvec{A}{AB}$ at discrete time steps. \ourlib{} follows the principle that ``everything which is not forbidden is allowed'', and never requires distinct descriptors.\footnote{In fact, all rules in~\cref{tab:rules} are satisfied by expressions with all-identical (or all-unset) descriptors, such as $\fvecg{A}{AA}$, and using semantics checking in \ourlib{} is entirely optional.} Our choice of rules may change in future work based on use of the library, as it is intended to support best practices without encumbering common tasks.

For example, it is common to update a rotation by integrating an angular velocity at discrete time steps:
\begin{equation} \label{eq:integrate}
\frot{AB} \coloneqq \frot{AB} \boxplus (\Delta t \cdot\f\omega{A}{AB})^\hatop.
\end{equation}
While the left hand side represents a perturbed or updated version of $\frot{AB}$, explicitly representing the perturbed frame would require $\boxplus$ to introduce a coordinate frame label not present in its operands. We do not force the declaration of a  perturbed frame, resulting in the rule~\cref{eq:rule-boxplus} for~$\boxplus$.

Starting with~\cref{eq:rule-boxplus}, we apply~\cref{eq:boxplus} to obtain
\begin{equation}
\fgrp{AB} = \exp(\falg{A}{AB})\circ\fgrp{AB}.
\end{equation}
It follows that \cref{eq:rule-expmap} is $\exp(\falg{A}{AB}) = \fgrp{AA}$. This makes the exponential map semantically non-bijective: it loses coordinate frame information, for which $\log(\exp(\valg)) = \valg$ does not hold. Consequently, applying \cref{eq:boxminus} to \cref{eq:rule-boxminus} produces the unsatisfactory result $\log(\fgrp{AA}) = \falg{A}{AB}$. A solution is to add an extra coordinate frame argument to the logarithmic map, as shown in \cref{eq:rule-logmap}.

\section{Results} \label{results}

We evaluate the runtime of our implementation compared to Ceres 1.13, GTSAM 4.0.0-alpha2, and hand-coded derivatives using Eigen 3.3.4.
Clang 5.0 was used with optimization flags \lstinline{-O3 -DNDEBUG -march=native} on an Intel Core i7 Skylake processor.
The Google Benchmark library was used for timing.
Results were averaged over repeated trials on sequences of random rotations.

\subsection{Rotation Chain} \label{res-chain}

First, we consider an increasingly long chain of rotations
\begin{equation} \label{eq:rotate_chain}
\fvec{}{2} = \left(\prod_{i=1}^{N} \frot{i}\right) \fvec{}{1}
\end{equation}
where $\vrot_i$ are rotation matrices. For example, for $N=3$,
\begin{equation} \label{eq:rotate_chain2}
\fvec{}{2} = \frot{1} \frot{2} \frot{3} \fvec{}{1}.
\end{equation}
This example is similar to~\cite[eq. (6)]{Sommer2013}.
Applying the chain rule to the derivatives found in~\cite{Bloesch2016} gives
\begin{align}
\partial{\fvec{}{2}}/\partial{\fvec{}{1}} &= \frot{1} \frot{2} \frot{3}, \label{eq:J0} \\
\partial{\fvec{}{2}}/\partial{\frot{1}} &= -\fvec{}{2}^\hatop, \label{eq:J1}\\
\partial{\fvec{}{2}}/\partial{\frot{2}} &= -\fvec{}{2}^\hatop\frot{1}, \label{eq:J2}\\
\partial{\fvec{}{2}}/\partial{\frot{3}} &= -\fvec{}{2}^\hatop\frot{1}\frot{2} \label{eq:J3}.
\end{align}

We use Eigen to hand-code \cref{eq:rotate_chain2,eq:J0,eq:J1,eq:J2,eq:J3}, reusing intermediate values and evaluating $-\vvec^\hatop \vrot$ as column-wise cross products for efficiency.

Using Ceres presents a challenge, as explained with example code in~\cite{Sommer2013}: its \texttt{AutoDiffCostFunction} produces global Jacobians. Obtaining a local Jacobian requires the extra calculation of the derivative of the global parametrization with respect to the local. This is so inefficient for rotation matrices that it is not a realistic use case of Ceres, and the results we show for Ceres use quaternions.

Using GTSAM, we differentiate \cref{eq:rotate_chain2} as shown in \cref{lst:gtsam_code}.

\begin{figure*}
\subfloat[]{
  \includegraphics{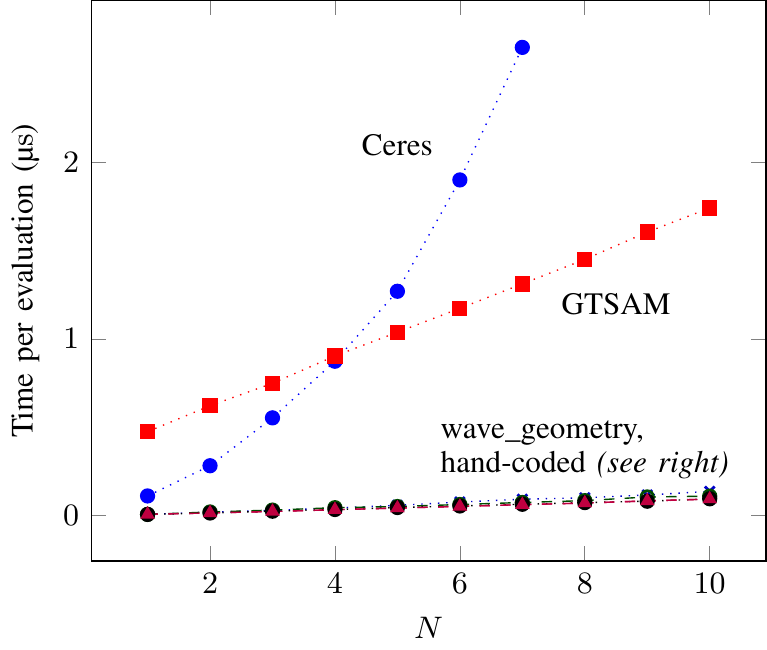}
}
\label{1a}
\subfloat[]{
  \includegraphics{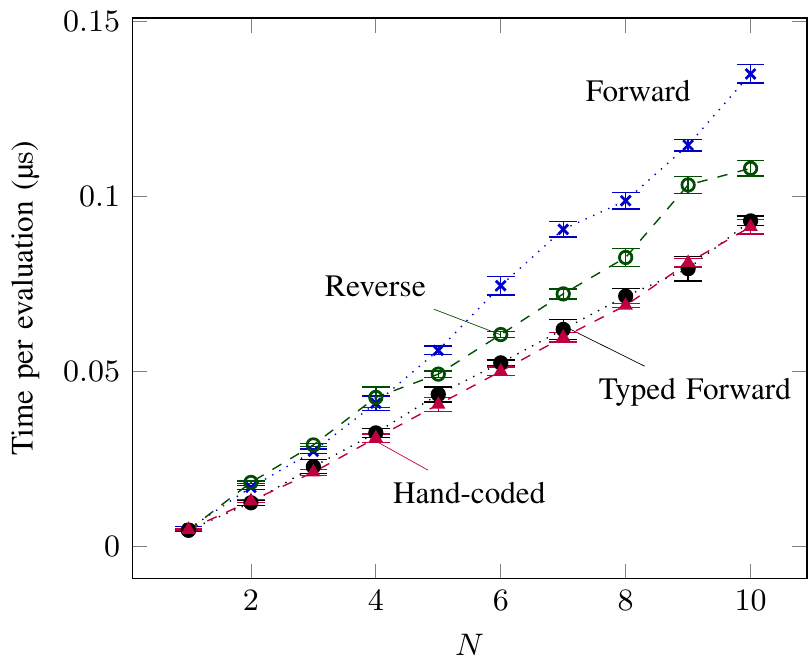}
}
\label{1b}
\caption{Comparison of time taken to evaluate result and all $N+1$ Jacobians in a chain of $N$ rotations \cref{eq:rotate_chain}. Results are averaged over many trials. (a) compares our results to existing libraries. (b) shows the same data at a larger scale, comparing our three implementations to the hand-coded reference.}
\label{fig:rotate1}
\end{figure*}

\begin{lstlisting}[caption={Automatic differentiation of a rotation chain in GTSAM}, label={lst:gtsam_code}]
// Define expressions for inputs
Expression<Rot3> R1_{'R', 1}, R2_{'R', 2}, R3_{'R', 3};
Expression<Point3> p1_{'p', 1};

// Define expression for the rotation chain
Expression<Point3> p2_ = rotate(R1_ * R2_ * R3_, p1_);

// For each symbol, set a linearization point
Values values{};
values.insert(Symbol{'R', 1}, getRotation());
values.insert(Symbol{'R', 2}, getRotation()});
values.insert(Symbol{'R', 3}, getRotation()});

// Get result and all Jacobians
std::vector<Matrix> jacobians(4);
Point3 p2 = p2_.value(values, jacobians);
\end{lstlisting}

\Cref{lst:wave_forward} demonstrates the same task in \ourlib{}, using the forward evaluator combined with coordinate frame semantics (labelled ``typed forward'' in results).
The reverse evaluator is invoked by passing no arguments, as demonstrated in \cref{lst:wave_reverse}.
While these examples show C++17 syntax, the library can be used in C++11 and above.

\begin{lstlisting}[caption={Automatic differentiation of a rotation chain in \ourlib{}}, label={lst:wave_forward}]
// Define inputs (with frame semantics)
wave::RotationMd<D, C> R1 = getRotation();
wave::RotationMd<C, B> R2 = getRotation();
wave::RotationMd<B, A> R3 = getRotation();
wave::Pointd<A> p1 = getPoint();

// Define the expression and differentiate
auto expr = R1 * R2 * R3 * p1;
auto[p2, J1, J2, J3, Jp] = expr.evalWithJacobians(R1, R2, R3, p1);
\end{lstlisting}
\begin{lstlisting}[caption={Reverse-mode AD in \ourlib{}}, label={lst:wave_reverse}]
// Differentiate in reverse mode
auto[p2, J1, J2, J3, Jp] =
   (R1 * R2 * R3 * p1).evalWithJacobians();
\end{lstlisting}

\Cref{fig:rotate1} presents the results for $N$ from 1 to 10. For this function, all three \ourlib{} methods clearly outperform the existing libraries. The time taken by Ceres grows rapidly with $N$, matching the results of~\cite{Sommer2013}. GTSAM has a high initial overhead, but scales linearly, at a rate  about 14 times that of the hand-coded reference.

Our typed forward evaluator's performance matches the reference, while the reverse evaluator has an average overhead of 24\%. This represents an improvement over the~4$\times$ slowdown reported in~\cite{Sommer2013} for a similar example with quaternions. The typed forward evaluator can outperform the reverse because it naturally exploits the structure of this problem. In the next example, that is not the case.

GTSAM is disadvantaged in this comparison because it is designed for calculating sparse Jacobians of large graphs, not individual expressions. While our approach is faster than GTSAM's runtime tree, it does have a limitation: it requires advance knowledge of function flow, and cannot be used on arbitrary functions with unpredictable branching and loops, or on expressions composed at runtime. A combination of the two methods, using optimized ET-based AD within subtrees of a larger graph, could be a valuable improvement.

\subsection{IMU Factor} \label{res-imu}
Next, we evaluate our work on a sample expression simplified from a preintegrated IMU factor~\cite[eq. (45)]{forster2017manifold}.
Let $\v {\tilde{C}}_{IJ}$ represent a preintegrated measurement of rotation between times $i$ and $j$, for which $\frot{WI}$ and $\frot{WJ}$ are the estimated orientations in the world frame. Let $\varphi$ be an unknown small change in bias. The residual of the bias-updated preintegrated measurement is
\begin{equation}  \label{eq:imuexample}
\v r_{ij} = \left(\v {\tilde{C}}_{IJ} \boxplus \v \varphi\right) \boxminus \frot{WI}^{-1}\frot{WJ}
\end{equation}
which can be expressed as
\begin{equation} \label{eq:imuexample2}
\v r_{ij} = \log\left(\left(\v{\tilde{C}}_{IJ} \exp(\v \varphi)\right)^{-1} \circ \frot{WI}^{-1}\frot{WJ}\right).
\end{equation}
Each of the four Jacobians of \cref{eq:imuexample2} contains the derivative of the logarithmic map~\cite[eq. (9)]{forster2017manifold} which, compared to the derivatives of  \cref{eq:rotate_chain}, is expensive to compute.

\Cref{tab:imuresults} presents the results. As expected when multiple Jacobians rely on an intermediate Jacobian calculation, the reverse evaluator outperforms the forward evaluator in this example, and is approximately 20\%  slower than the hand-coded reference.

\begin{table}[tb]
\caption{Time to Evaluate Value and Jacobians of \cref{eq:imuexample}} \label{tab:imuresults}
\setlength{\tabcolsep}{1pt} 
\begin{tabu} to \columnwidth { l *{4}{X[c]} }
\toprule 
Algorithm & Hand-coded & Forward & Typed Forward & Reverse \\
\midrule
Mean (ns) & \num{302} & \num{642} & \num{596} & \num{360} \\
Std. dev. (ns) & \num{10} & \num{29} & \num{27} & \num{18} \\
\bottomrule
\end{tabu}
\end{table}

\addtolength{\textheight}{-0.0cm}   

\section{Conclusion}

This paper introduces \ourlib{}, a C++ library for robotics aimed at performance and rapid development. It incorporates a method for fast forward- and reverse-mode automatic differentiation of geometric expressions on manifolds. For representative examples on $\SO3$, the reverse-mode evaluator achieves an average overhead of 20--25\% over hand-coded analytic derivatives. While hand-optimized derivatives will always have a place in performance-critical code, such as in embedded and high-volume production systems, these results are promising for rapid prototyping and research.

We also present a system for coordinate frame semantics checking that catches common errors at compile time. Because it encodes additional type information into each expression, this system complements ET-based AD.

Our AD method outperforms existing libraries such as GTSAM by applying knowledge to expression trees at compile time, and is currently limited to closed-form expressions. Future work includes support for runtime composition of arbitrary functions, tight integration with a nonlinear least squares optimizer, and the use of cost estimates for selection of AD algorithms and tree transformations.


\bibliographystyle{IEEEtran}
\bibliography{IEEEabrv,geometry}


\end{document}